\documentclass[10pt,twocolumn,letterpaper]{article}

\usepackage{cvpr}
\usepackage{times}
\usepackage{epsfig}
\usepackage{graphicx}
\usepackage{amsmath}
\usepackage{amssymb}
\let\emptyset\varnothing
\usepackage{bm}
\usepackage{algorithm}
\usepackage{algorithmic}

\usepackage[pagebackref=true,breaklinks=true,letterpaper=true,colorlinks,bookmarks=false]{hyperref}

\cvprfinalcopy 

\ifcvprfinal\fi

\begin{document}

\title{Selective Synthetic Augmentation with Quality Assurance}

\author{Yuan Xue,\textsuperscript{\rm 1}\thanks{Authors contributed equally}
\hspace{1.3mm}
Jiarong Ye,\textsuperscript{\rm 1}${}^{*}$
\hspace{0.8mm}
Rodney Long,\textsuperscript{\rm 2}
\hspace{0.8mm}
Sameer Antani,\textsuperscript{\rm 2}
\hspace{0.8mm}
Zhiyun Xue,\textsuperscript{\rm 2}
\hspace{0.8mm}
Xiaolei Huang\textsuperscript{\rm 1}\\
${}^1$The Pennsylvania State University \qquad\qquad ${}^2$National Institutes of Health
}

\maketitle

\begin{abstract}
Supervised training of an automated medical image analysis system often requires a large amount of expert annotations that are hard to collect. Moreover, the proportions of data available across different classes may be highly imbalanced for rare diseases. To mitigate these issues, we investigate a novel data augmentation pipeline that selectively adds new synthetic images generated by conditional Adversarial Networks (cGANs), rather than extending directly the training set with synthetic images. The selection mechanisms that we introduce to the synthetic augmentation pipeline are motivated by the observation that, although cGAN-generated images can be visually appealing, they are not guaranteed to contain essential features for classification performance improvement. By selecting synthetic images based on the confidence of their assigned labels and their feature similarity to real labeled images, our framework provides quality assurance to synthetic augmentation by ensuring that adding the selected synthetic images to the training set will improve performance. We evaluate our model on a medical histopathology dataset, and two natural image classification benchmarks, CIFAR10 and SVHN.
Results on these datasets show significant and consistent improvements in classification performance (with $\textbf{6.8}\%, \textbf{3.9}\%, \textbf{1.6}\%$ higher accuracy, respectively) by leveraging cGAN generated images with selective augmentation.
\end{abstract}

\section{Introduction}
The development of image recognition systems has been greatly advanced by deep learning in the past decade.
Towards medical image recognition, recent results have inspired hope that automated or computer-assisted systems can effectively reduce clinicians' workload and improve the quality of care in underdeveloped regions of the world.
For example, an automatic cervical intraepithelial neoplasia (CIN) grading system for cervical histopathology images may help pathologists with diagnosis and also potentially mitigate the problem of inter- and intra- pathologist variation in disease assessment.

The supervised training of image recognition systems often requires huge amounts of expert annotated data to reach a high level of accuracy. However, for many practical applications especially in the medical domain, only small datasets of labeled data are available due to annotation cost and privacy concerns, and the labels are often imbalanced between disease and healthy classes.
While traditional data augmentation can increase the amount of training data to some degree, the methods usually employed (such as cropping and flipping) lack flexibility and cannot fill the whole data distribution with missing data samples. 

Motivated by the difficulties discussed above in creating sufficiently large training sets for medical image recognition systems, we focus on the problem of expanding training sets with high-quality synthetic examples.
Recently, several works in medical image analysis have leveraged unsupervised learning methods, more specifically, Generative Adversarial Networks (GANs)~\cite{goodfellow2014generative}, to mitigate the effects of small training sets on network training
~\cite{liu2019wasserstein,frid2018gan}. These works show that carefully designed GANs can generate visually appealing synthetic images, but two major problems remain unsolved for generalized and robust synthetic augmentation: (1)
how to measure label uncertainty of generated images; and (2) how to ensure the feature quality of synthetic images used for data augmentation.
In other words, blindly adding synthetic samples to the original training set, even if they are visually realistic, is not guaranteed to improve the classification model performance, and can potentially adversely alter the data distribution and downgrade model performance. We provide a detailed analysis in Section~\ref{sec:results}.

In this paper, we aim at solving these two problems by  \textit{selectively} adding synthetic samples generated by conditional GANs to the original training set. The whole selective synthetic augmentation pipeline consists of two steps: 1)
cGAN training with model selection in which a smoothed version of FID score is used to select the best cGAN model, and 2) synthetic image selection based on the average entropy of a generated image calculated from multiple runs of a feature extractor with Monte Carlo dropout (MC-dropout) and the divergence in feature space between the synthetic image and real-data class centroids.
More specifically, for synthetic image selection, 
we pre-train a feature extractor with MC-dropout~\cite{gal2016dropout}  using a ResNet34~\cite{he2016deep} model to calculate feature centroids for each class as the average points of all training images in the feature space. To select generated samples which are matched to their assigned (i.e. conditional) labels with high confidence, we ran the pre-trained feature extractor multiple times with Monte Carlo sampling on the synthetic images. We first calculate the expectation of predictive entropy for each sample and keep those samples with relatively low entropy (i.e. high label confidence) for the next step. 
We then calculate the mean centroid distances from Monte Carlo sampling and keep samples with relatively small distances to centroids in feature space. The total number of selected samples is decided according to the augmentation ratio $r$ (i.e., the proportion of the number of augmented samples to the number of original training samples). Experimental results show that our proposed cGAN model along with  feature-based filtering significantly outperforms the baseline ResNet34~\cite{he2016deep} model with traditional augmentation, and also outperforms the synthetic augmentation methods without filtering.

To validate the effectiveness and generality of our proposed selective synthetic augmentation pipeline, we conduct extensive experiments on both medical images and natural images. We first study the 4-class (Normal, CIN 1-3) cervical histopathology image classification problem and evaluate our models on a heterogeneous epithelium image dataset~\cite{xue2019synthetic} with  limited and highly unbalanced numbers of patch-level annotations per class label.
We design a new cGAN model to synthesize high-fidelity epithelium histopathology patches to expand the training data. To prove the robustness and general applicability of this model, we experiment on two well-known natural image classification datasets: CIFAR10~\cite{krizhevsky2009learning} and SVHN~\cite{netzer2011reading}. To simulate the scenario of insufficient amount of training data, we choose to train our model on different training subsets of CIFAR10 with various sizes and a small subset of the SVHN dataset. We compare our proposed selective synthetic augmentation method with baseline methods including baseline classification models, models trained with traditional augmentation, and models trained with synthetic augmentation but without selective filtering. Experimental results show that our model achieves significant improvements with $\textbf{6.8}\%, \textbf{3.9}\%, \textbf{1.6}\%$ higher accuracy than baseline classification models on histopathology, CIFAR10 and SVHN datasets, respectively.

The main contributions of this work are as follows:
\begin{itemize}
\setlength{\itemsep}{-0.4em}
  \item We design a novel conditional GAN model for synthesizing photo-realistic histopathology patches.  A smoothed version of FID score  is proposed as a metric for model selection and is used to select the best cGAN model. With a limited amount of training data, our synthetic images achieve both high fidelity and large diversity.
  \item We propose a selective synthetic augmentation method that aims at elevating the performance of training classification systems with limited amounts of data or highly imbalanced data. The proposed method is general and can be employed on multiple applications.
  \item We conduct extensive experiments on one medical image dataset and two natural image datasets. Compared with baseline models, including a state-of-the-art synthetic augmentation model, our proposed method improves the augmented classification performance by a large margin.
\end{itemize}

\section{Related Work}
\subsection{Conditional Image Synthesis}
Generative adversarial networks (GANs)~\cite{goodfellow2014generative} as an unsupervised learning technique, has enabled a wide variety
of applications including image synthesis, object detection~\cite{li2017perceptual} and image segmentation~\cite{xue2018segan}. Among variants of GANs, conditional GAN (cGAN) generates~\cite{mirza2014conditional, odena2017conditional} more interpretable results with conditional inputs. For instance, images can be generated conditioning on class labels, which enables cGAN to serve as a tool to generate labeled samples for synthetic augmentation. Current state-of-the-art cGAN models often contain progressive refinement procedures~\cite{zhang2018stackgan++, karras2017progressive} or large scale training~\cite{brock2018large}, which enable them to generate high fidelity images. In this work, we use our proposed cGAN and state-of-the-art BigGAN~\cite{brock2018large} to generate synthetic images to augment classification models.

\subsection{Synthetic Data Augmentation}
To better utilize training data and reduce over-fitting during the training process, data augmentation has become a common practice for training deep neural networks. Traditional data augmentation~\cite{perez2017effectiveness} often involves transformations applied directly on original training data, such as cropping, flipping and color jittering. While serving as an implicit regularization, straightforward data augmentation techniques are limited in augmentation size and diversity. Moreover, they tend to generate plausible images which can disrupt the original data distribution. To overcome the limitation of traditional augmentation, several works have been done to improve the effectiveness of data augmentation. Rather than using pre-defined augmentation policy, Auto Augmentations~\cite{cubuk2019autoaugment, ho2019population} use hyper-parameter searching to automatically find the optimal augmentation policy. Another popular trend is to generate synthetic images to increase the amount and diversity of original dataset, for which we denote as Synthetic Augmentation. Along this direction, Ratner~\emph{et al.}~\cite{ratner2017learning} learns data transformation with unlabeled data using GANs. GAGAN~\cite{antoniou2017data} and BAGAN~\cite{mariani2018bagan} uses cGANs~\cite{mirza2014conditional} generated samples to augment the standard classifier in the low-data regime. 

Compared with works done in the natural image domain, the insufficient and imbalanced data issue is more prominent in medical image applications. To mitigate there problems, researchers have been working on synthetic augmentation for medical image recognition tasks. 
Frid-Adar~\emph{et al.}~\cite{frid2018gan} proposes to use cGAN generated synthetic CT images to improve the performance of CNN in liver lesion classification. Gupta~\emph{et al.}~\cite{gupta2019generative} synthesizes lesion images from non-lesion ones using CycleGAN~\cite{zhu2017unpaired}. Bowles~\emph{et al.}~\cite{bowles2018gan} uses GAN derived synthetic images to augment medical image segmentation models. Although pioneer works on GAN based synthetic augmentation have achieved promising results, most of them choose to blindly add synthetic samples to the original data and few work considers how to assure the quality of synthetic images or controlling the augmentation step after image synthesis. Very recently, Xue~\emph{et al.}~\cite{xue2019synthetic} proposes a feature based filtering mechanism for synthetic augmentation. While improving the classification performance, their GAN generated results are not realistic enough and their method is hard to generalize to other tasks or modalities. In this work, we aim at developing a whole synthetic augmentation pipeline by selectively adding generated samples to the original dataset. Compared with previous works, our proposed selective synthetic augmentation is general and applicable to various tasks.

\section{Methodology}

\begin{figure*}
\begin{center}
  \includegraphics[width=0.80\linewidth]{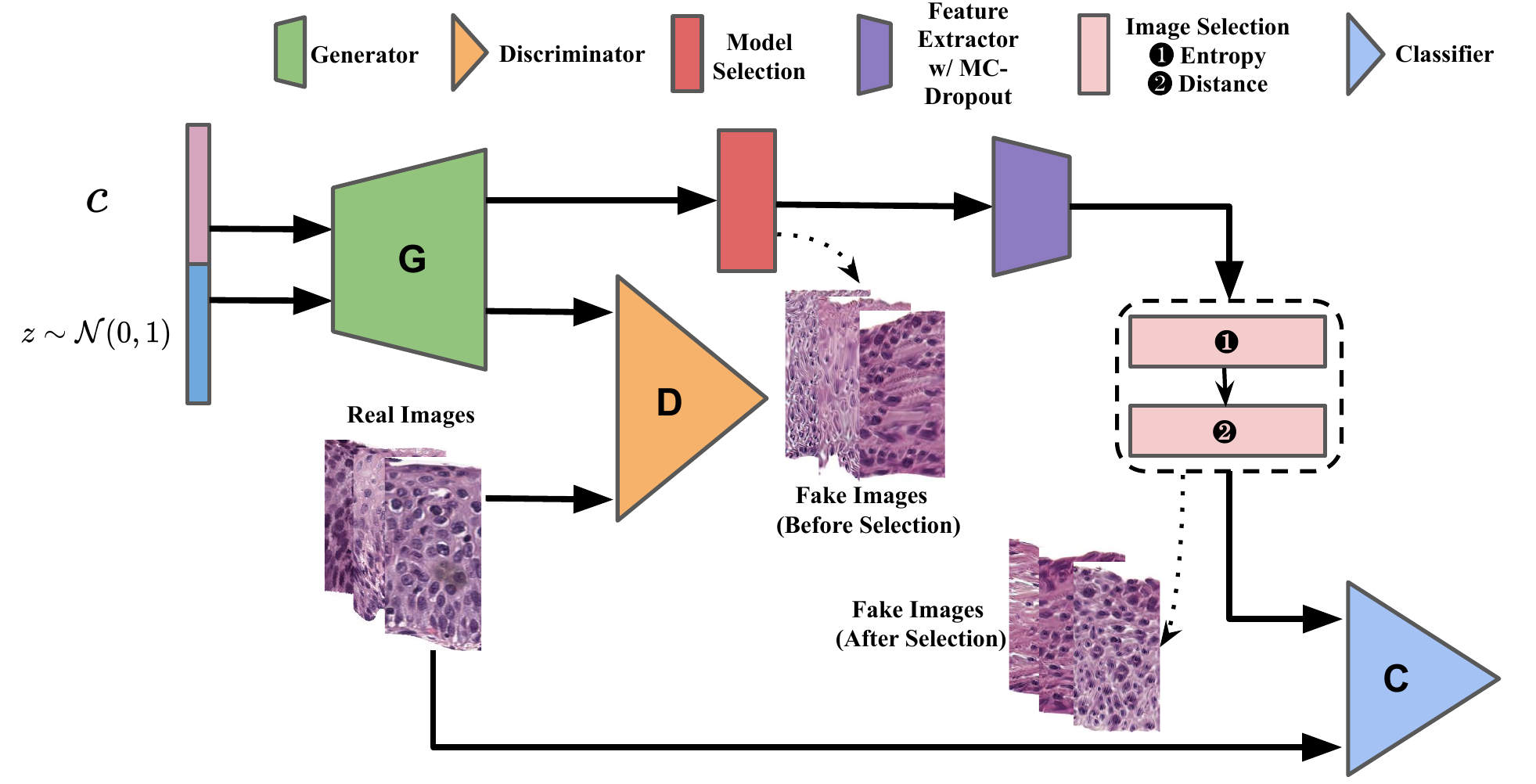}
\end{center}
  \caption{The architecture of the proposed selective synthetic augmentation pipeline.}
\label{fig:architecture}
\end{figure*}

An overall illustration of our proposed data augmentation pipeline can be found in Fig.~\ref{fig:architecture}. We first train a conditional GAN model based on the labeled training images. The best model is selected based on our proposed smoothed FID score~\cite{heusel2017gans}. A pool of synthetic images are then generated using the selected model. All images are then passed into the image selection module to filter out the ones that fail to contribute sufficient amount of meaningful information. After image selection, a classification model is trained with both original and synthetic training data. Trained classification models can then be used for inference on test data. More details are introduced in the following subsections. 

\subsection{Conditional GAN models}

The conventional cGANs~\cite{mirza2014conditional} have an objective function defined as:
\begin{small}
\begin{multline}
\min_{\theta_G} \max_{\theta_D} \mathcal{L}_{\text{cGAN}} = \mathbb{E}_{x\sim p_\mathrm{data}}[\log D(x,c)] + \\ \mathbb{E}_{z\sim \mathcal{N}}[\log (1 - D(G(z,c)))] \enspace .\label{Eq:cGAN}
\end{multline}
\end{small}
In the equation above, $x$ represents the real data from an unknown image distribution $p_\mathrm{data}$ and $c$ is the conditional label (\emph{e.g.}, CIN grades). $z$ is a random vector for the generator $G$, drawn from a standard normal distribution $\mathcal{N}(0,1)$. During the training, $G$ and $D$ are alternatively optimized to compete with each other.

Since there are no existing cGAN framework specified for cervical histopathology images, we choose to design a new model based on the state-of-the-art StackGAN V2~\cite{zhang2018stackgan++} model. StackGAN generates synthetic images in a corase-to-fine fashion through multiple stages, where details of images are gradually refined to guarantee the fidelity. The training procedure of StackGAN is similar to Eq.~\ref{Eq:cGAN}. The generator of stage I takes a random noise and labels as input, and the generator of stage II and III take the output of the previous stage as input instead of random noise. We also incorporate the minibatch discrimination module~\cite{salimans2016improved} into our discriminator to 
increase diversity among the generated examples and mitigate the issue of mode collapse indicated by the high homogeneity of the synthetic image pool.
Following state-of-the-art works in conditional image synthesis~\cite{zhang2019self,brock2018large}, spectral normalization~\cite{miyato2018spectral} is utilized in discriminators of all stages to improve model performance.

\begin{figure}[b]
\begin{center}
  \includegraphics[width=0.99\linewidth]{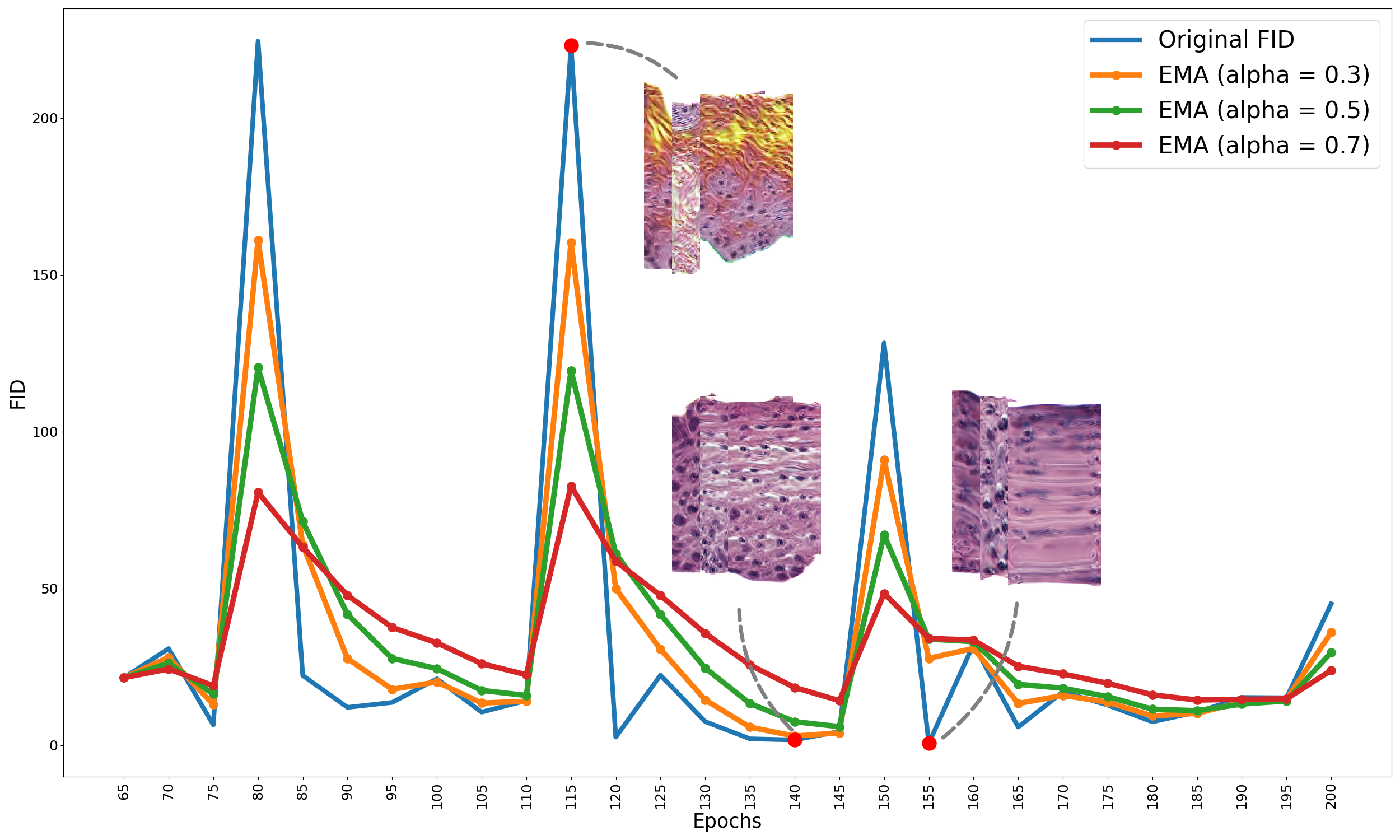}
\end{center}
  \caption{FID scores (pre-trained ResNet34) of saved cGAN models. Scores are smoothed with varying EMA parameter $\alpha$.}
\label{fig:fid}
\vspace{-10pt}
\end{figure}

The main architecture of our cGAN model specifically designed for cervical histopathology images is similar to StackGAN~\cite{zhang2018stackgan++}, with WGAN-GP applied on the loss of the discriminators of all 3 stages. The input of the stage I generator is the concatenation of random noise and class labels (CIN1-3, Normal) that are first one-hot encoded and then embedded by a transposed convolution layer. The stage I generator consists of $4$ up-sampling blocks with $3\times3$ conv kernels, the layers in each block are of the same structure but different in and out channels. 
Two other blocks of the same architecture but different in and out channels are employed in stage II and III generator respectively, after a set of residual blocks. 
With $3$-scale generator and discriminator pairs, we have images of increasing resolution across $3$ scales generated. 
Inside each discriminator, the main structure contains several down-sampling layers with $4\times4$ conv kernels. 
The down-sampling layers are followed by a $3\times3$ conv layer, a spectral normalization layer, a batch normalization layer, a Leaky ReLU activation layer, a minibatch discrimination~\cite{salimans2016improved} block for preventing mode collapse during GAN training, and a fully connected layer for the final output.

Compared with medical data such as the cervical histopathology images, natural images tend to have larger discrepancy between image classes. To ensure both the image quality and diversity, we use the state-of-the-art large scale BigGAN~\cite{brock2018large} model to generate photo-realistic images for natural images. After generating enough number of synthetic images, we apply the same image selection mechanism as in medical images.

\subsection{Model Selection}\label{model_selection}

The first step of our selective synthetic augmentation pipeline is to choose the trained cGAN model for image synthesis. An ideal solution is using a metric to simulate the human evaluation process for model selection, however, finding a proper evaluation metric for GAN models or image synthesis results remains to be an unsolved problem. For natural image synthesis tasks, Inception Score~\cite{salimans2016improved} and FID~\cite{heusel2017gans} score are two commonly used metrics. The calculation of these two metrics rely on the pre-trained Inception V3~\cite{szegedy2016rethinking} model trained on ImageNet~\cite{deng2009imagenet}. However, the distribution of natural images and medical images such as cervical histopathology images can be quite different. We follow the calculation of original FID score while replacing the Inception V3 model with a ResNet34~\cite{he2016deep} model pre-trained on the same dataset. 

Although the FID score itself cannot guarantee agreement with human judgment, trends of FID often provide a reliable estimation of the quality of a GAN model. However, as we can observe from Figure~\ref{fig:fid}, due to the instability in GAN training, the FID scores of each saved epoch fluctuate constantly and fails to provide a distinguishable pattern. Based on the trend of the unadjusted FID score, we should choose the model saved at epoch 155. However, one can see that images generated by the chosen model is not satisfactory in Figure~\ref{fig:fid}. To get a robust estimation of model quality and mitigate the effect caused by the deviated outliers, we apply the Exponential Moving Average (EMA)~\cite{hunter1986exponentially} algorithm for smoothing the original FID score as:

\begin{small}
\begin{equation}
    \hat{d} = \left\{\begin{matrix} \text{FID}_1,& \enspace t=1
\\ \alpha \hat{d}_{t-1} + (1-\alpha)\text{FID}_t,& \enspace t>1 \label{Eq:smoothFID}
\end{matrix}\right.
\end{equation}
\end{small}

We monitor the training process with smoothed FID. As shown in~\ref{fig:fid}, different values of $\alpha$ lead to similar trend and the chosen model associated with lowest smoothed FID has better image quality than the model chosen by original FID score. 
$\alpha$ is set to $0.3$ for the optimal pattern display. After smoothing with EMA algorithm, the global minimum is reached at epoch 140, which is the GAN model we saved for further data augmentation.

\begin{figure}[!ht]
\begin{center}
  \includegraphics[width=0.99\linewidth]{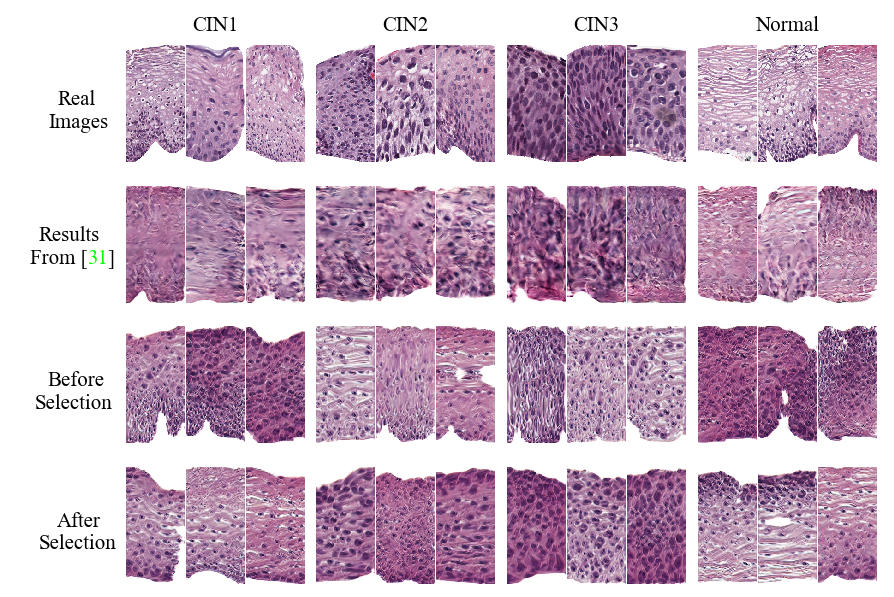}
\end{center}
  \caption{Examples of real images, synthetic images generated from~\cite{xue2019synthetic}, and images generated by our cGAN model trained on histopathology dataset before and after selection. Our cGAN generates realistic images with clearly better visual quality than~\cite{xue2019synthetic}. Zoom in for better view.}
\label{fig:cervical}
\end{figure}

\begin{figure*}[!ht]
\begin{center}
  \includegraphics[width=0.95\linewidth]{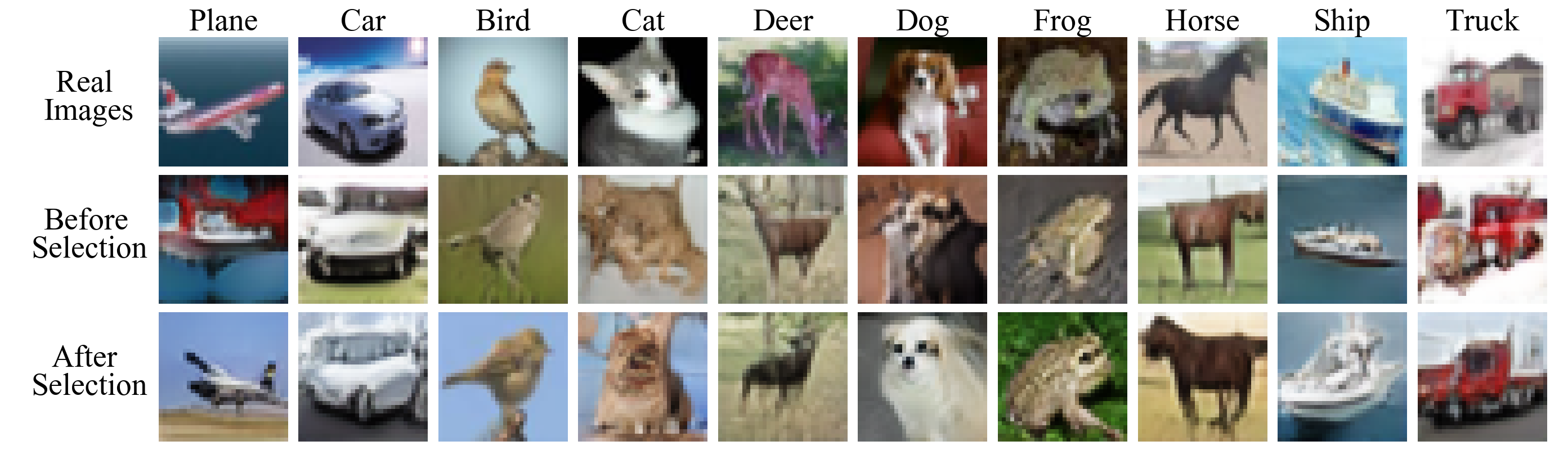}
\end{center}
  \caption{Examples of real and synthetic images generated by BigGAN~\cite{brock2018large} trained on 10\% of CIFAR dataset.}
\label{fig:cifar}
\end{figure*}

\subsection{Image Selection}\label{img_selection}
Given a trained cGAN model, one can sample infinite number of inputs from the Gaussian distribution and generate infinite number of synthetic images. While a good cGAN model can guarantee that generated images are realistic enough, there are no guarantee that those images can be used to augment the original training set for visual recognition tasks. In current GAN-based data augmentation works, with different data augmentation ratio, different number of generated images are added to the training set. However, the effectiveness of such augmentation pipeline is heavily affected by varying quality of synthetic images as well as the diversity of the images. To reduce the randomness in the synthetic augmentation process and selectively add in new images, we break the whole process into two steps: find samples that can be confidently match with the label assigned to them during the conditional generation; then find samples that are within the certain neighborhood of class centroids in the feature space to provide meaningful features. Such steps are done with a pre-trained feature extractor to calculate centroids for real samples and extract features for fake samples. Considering that a single feature extractor cannot provide robust feature extraction results, we use a feature extractor with Monte Carlo dropout (MC-dropout)~\cite{gal2016dropout} and take the expectation value of multiple samplings to reduce the uncertainty of feature extraction.

In the first step, we calculate entropy score between the probability distribution of fake samples. If the feature extractor is confident that samples can be classified into certain class, the entropy score would be low. We rank the entropy score of all generated images in ascending order and choose the first half with lower entropy. We assume that this step can filter out considerable amount of samples with indistinguishable features among classes. The necessity of entropy selection is proved by experiments on the cervical dataset, which will later be discuss in Section~\ref{Experiments}.

\begin{algorithm}[!t]
\begin{algorithmic}

    \caption{Selective Synthetic Augmentation}\label{algorithm} 
    \STATE \textbf{Input}: a set of saved cGAN models \{$G_n$\}, number of classes $\mathcal{C}$, augmentation ratio $r$, number of original training samples $N = \sum_{i=1}^{\mathcal{C}} N_i$.
    \STATE \textbf{Output}: selected synthetic samples $\mathcal{X}$ with $|\mathcal{X}|=rN$. 
    \STATE \textbf{Initialization}: $\mathcal{X}_1 = \emptyset$,
    $\hat{n} = \arg \min (\hat{d_n})$,
    $G_{\hat{n}}$ generated samples $\mathcal{X}_{0}$ = \{$x_j^i: i \leq \mathcal{C}, j \leq 4rN_i$ \},
    entropy $\mathcal{E}^i = \{e_j^i: e_j^i = - \sum p_j^i \log p_j^i, i \leq \mathcal{C}, j \leq 4rN_i \}$. \\
    \FOR {$x_j^i \in \mathcal{X}_0$}
    \IF {$e_j^i < \text{Median} (\mathcal{E}^i)$}
    \STATE $\mathcal{X}_1 = \mathcal{X}_1 \cup \{x_j^i\}$ 
    \ENDIF
    \ENDFOR \\
    class centroid distance $\mathcal{D}^i = \{d_j^i: d_j^i = D_f (x_j^i, c_i)\}$. \\
    \FOR {$x_j^i \in \mathcal{X}_1$} 
    \STATE $d_j^i = D_f \{x_j^i, c_i\}$ 
    \IF{$d_j^i < \text{Median} (\mathcal{D}^i)$}
    \STATE $\mathcal{X} = \mathcal{X} \cup \{x_j^i\}$ 
    \ENDIF
    \ENDFOR
\end{algorithmic}
\end{algorithm}

After the entropy selection, we further select synthetic images based on their distance to class centroids in the feature space. Feature extractors are also run multiple times with MC sampling and the average feature distances are calculated. Similar to~\cite{xue2019synthetic}, the feature distance between image $x$ and centroid $c$ is defined as 

\begin{small}
\begin{equation}
D_{f}(x,c_i) = \frac{1}{K}\sum_{k}\sum_{l} \frac{1}{H_lW_l} ||\hat{\phi_l^k}(x) - \hat{\phi_l^k}(c_i)||_{2}^{2} \enspace ,\label{Eq:feature}
\end{equation}
\end{small}

\noindent where $\hat{\phi_l^k}$ is the unit-normalized activation in
the channel dimension $A_l$ of the $l$th layer of the $k$-th MC sampling feature extraction network with shape $H_l \times W_l$. We denote the total sampling time as $K$. $D_{f}(x,c_i)$ can be regarded as an estimated cosine distance between sample and $i$-th centroid in the feature space.

The centroid $c$ is calculated as the average feature of all labeled training images in the same class. For class $i$, its centroid $c_i$ is represented by
\begin{small}
\begin{equation}
c_i = [\frac{1}{N_i}\sum_{j=1}^{N_i}\phi_1(x_j),...,\frac{1}{N_i}\sum_{j=1}^{N_i}\phi_L(x_j)] \enspace ,\label{Eq:centroid}
\end{equation}
\end{small}
\noindent where $N_i$ denotes the number of training samples in $i$th class and $x_j$ is the $j$th training sample. Similar to Eq.~\ref{Eq:feature}, $\phi_l$ is the activation extracted from the $l$th layer of the feature extraction network. $L$ is the total number of layers utilized in the feature distance selection. $c_i$ is retained by one time MC sampling and fixed during the distance calculation.

First half of remaining samples after entropy ranking with lower feature distances to its assigned centroids are kept in feature distance selection. We believe that this step can filter out considerable amount of samples whose assigned label do not represent their true label, so that these samples can be confidently added to the original training set.  
 
 In conclusion, given augmentation ratio $r$, we first generate $4rN_i$ images for each class, then select $rN_i$ images according to the process described above. As the synthetic augmentation performance highly relies on the quality and diversity of generated images, choosing $r$ with large value may bring in very similar samples and reduce the effectiveness of augmentation. We provide ablation study on $r$ in Section~\ref{sec:results}. A detailed description of our proposed selective synthetic augmentation pipeline is provided in Algorithm~\ref{algorithm}.

\section{Experiments}\label{Experiments}

\subsection{Datasets}
The first dataset contains labeled cervical histopathology images collected from a collaborating health sciences center. All images are annotated by the same pathologist. The data processing follows~\cite{xue2019synthetic}, results in patches with a unified size of $256 \times 128$. Compared with the dataset used in~\cite{xue2019synthetic}, we include more data for more comprehensive experiments. In total, there are $1,274$ Normal, $370$ CIN1, $541$ CIN2, $482$ CIN3 patches. Examples of the images can be found in the first row of Fig.~\ref{fig:cervical}. 
We randomly split the dataset, by patients, into training, validation, and test sets, with ratio 7:1:2 and keep the ratio of image classes almost the same among different sets. All evaluations and comparisons reported in this section are carried out on the test set.

To further prove the generality of our proposed method, We also conduct experiments on two commonly used natural image classification benchmarks: CIFAR10 and SVHN. To mimic the situation where only limited number of training data is available, we use different size of random subsets of CIFAR10 to train the cGAN model and the classifier. Models trained with different number of training samples are all evaluated on the full test set. For SVHN dataset, we also evaluate our method using randomly selected $25\%$ of the training set. 

\begin{figure*}[!ht]
\begin{center}
  \includegraphics[width=0.90\linewidth]{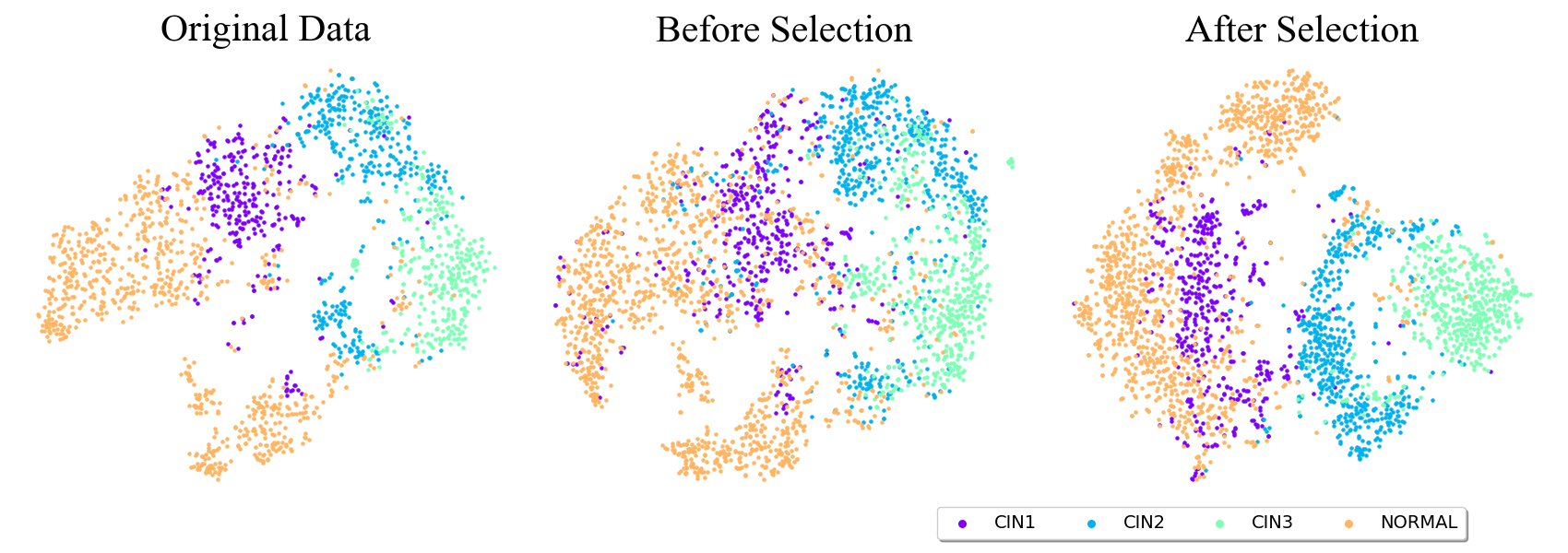}
\end{center}
  \caption{t-SNE of the ground truth and augmented histopathology training set before and after image selection.}
\label{fig:tsne_cervical}
\end{figure*}

\begin{figure*}[!ht]
\begin{center}
  \includegraphics[width=0.90\linewidth]{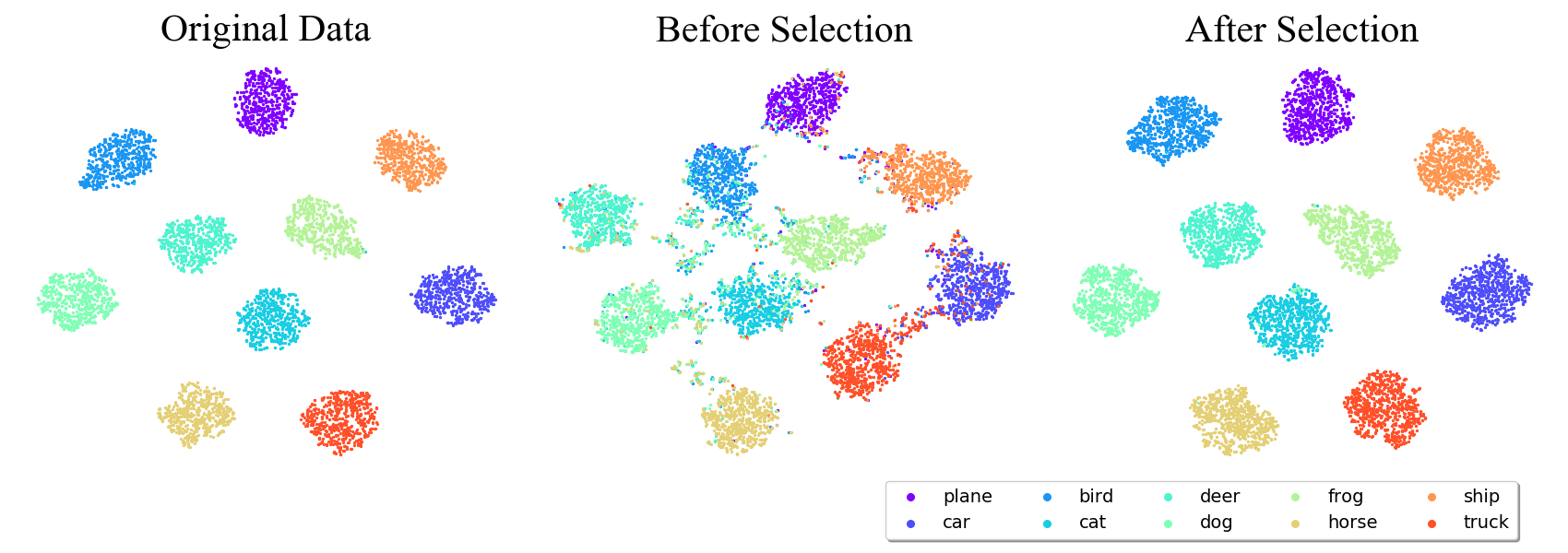}
\end{center}
  \caption{t-SNE of the ground truth and augmented CIFAR training subset(10\%) before and after image selection.}
\label{fig:tsne_cifar}
\end{figure*}

\subsection{Implementation Details}
The cGAN model used for generating cervical histopathology images is trained with batch size set to 64, fixed learning rate $5e-4$ and $200$ training epochs. Spectral normalization is used before batch normalization in discriminators of all stages, followed by Leaky ReLU activation and minibatch discrimination layers. For the two other natural image benchmark datasets, the state-of-the-art BigGAN~\cite{brock2018large} model is employed for image generation. The default hyperparameters of BigGAN specifically designed for training on CIFAR dataset are kept unchanged when extracting different proportion of subsets as the training set for GAN. We trained 500 epochs in total for each subset with batch size set to 50 and the learning rate of both generator and discriminator as 2e-4.

In the next step, GAN models are saved per 5 epochs for the model selection pipeline. For reasons mentioned in Section~\ref{model_selection}, the feature extractor used for FID score calculation in the model selection module is the same as our baseline classifier (ResNet34), followed by EMA-based smoothing to accentuate the pattern of synthetic image quality trend during the GAN training process. The GAN model selected for further stages of our purposed pipeline corresponds to the epoch with the lowest adjusted FID score. Next, we generate $4 rN_i$ synthetic images for each class $i$ with saved GAN, on which the same feature extractor is run for 5 times in order to extract the predicted probability from the softmax layer for entropy calculation, and also extract feature vectors after each residual block to obtain distance to centroids of ground truth. A dropout layer of rate $0.5$ is inserted before the last residual block right above the fully-connected layer of the feature extractor (ResNet34) for Monte Carlo sampling. Then the generated images are ranked based on the mean of entropy across 5 runs in ascending order, of which half images in each class are kept. The filtered pool of synthetic images are further ranked based on the mean of cosine distance to the centroid that corresponds to the assigned label of each image over 5 runs also in ascending order. Similarly, half are filtered out, leaving the rest for the final augmentation.

\subsection{Results Analysis} \label{sec:results}
To validate our proposed method, we conduct comprehensive experiments on both a cervical histopathology image dataset and two natural image datasets.
The image synthesis results for histopathology and CIFAR10 datasets are demonstrated in Fig.~\ref{fig:cervical} and Fig.~\ref{fig:cifar}, respectively. In both datasets, as we have already achieved promising image generation results, determining whether those samples can be used for data augmentation or not cannot be easily done by human observations. However, the discrepancy between images with and without selection is much more prominent in the feature space. After training a baseline ResNet34 classifier with the original training data, we use the pre-trained ResNet34 model as the feature extractor to extract features from the last convolutional layer in the ResNet model. We explore the distribution of training samples, including both original images and synthetic images, in the feature space using t-SNE~\cite{maaten2008visualizing}. In Fig.~\ref{fig:tsne_cervical}, without image selection, samples from different classes are
entangled together, introducing obscuring noise that disrupts the data distribution that real data presents. On the contrary, selected images have clearly more distinguishable features and can potentially help with improving the classification model performance. Similar phenomenon is also observed with more noticeable pattern in Fig.~\ref{fig:tsne_cifar}: while data augmentation without image selection increases the number of training samples, the original data distribution is distorted. After image selection, the original data distribution is recovered along with more number of data points.

\begin{table*}[t]
\begin{center}
\begin{tabular}{|l|c|c|c|c|}
\hline
{} &         Accuracy &              AUC &      Sensitivity &      Specificity \\
\hline
Baseline Model~\cite{he2016deep}        &  0.753 $\pm$ 0.012 &  0.835 $\pm$ 0.008 &  0.610 $\pm$ 0.025 &  0.889 $\pm$ 0.007 \\
\hline
\enspace + Traditional Augmentation &  0.756 $\pm$ 0.013 &  0.837 $\pm$ 0.008 &  0.598 $\pm$ 0.018 &  0.882 $\pm$ 0.008 \\
\hline
\enspace + GAN Augmentation, r=0.5      &  0.726 $\pm$ 0.007 &  0.817 $\pm$ 0.005 &  0.607 $\pm$ 0.008 &  0.871 $\pm$ 0.003 \\
\hline
\enspace + Single Filtering~\cite{xue2019synthetic}$^{*}$, r=0.5 &  0.804 $\pm$ 0.009 &  0.869 $\pm$ 0.006 &  0.696 $\pm$ 0.007 &  0.917 $\pm$ 0.006 \\
\hline
\enspace + Selective Augmentation, r=0.5   &  \textbf{0.821 $\pm$ 0.009} &  \textbf{0.881 $\pm$ 0.006} &  \textbf{0.706 $\pm$ 0.008} &  \textbf{0.927 $\pm$ 0.005} \\
\hline
\enspace + Selective Augmentation, r=0.8  &  0.802 $\pm$ 0.015 &  0.868 $\pm$ 0.010 &  0.686 $\pm$ 0.016 &  0.914 $\pm$ 0.007 \\
\hline
\enspace + Selective Augmentation, r=1.0  &  0.787 $\pm$ 0.021 &  0.858 $\pm$ 0.014 &  0.686 $\pm$ 0.021 &  0.912 $\pm$ 0.010 \\
\hline
\end{tabular}
\end{center}
\caption{Classification results of baseline and augmentation models with different settings. Each model is run 5 times for the calculation of all evaluation metrics. Note that we use improved GAN model and model selection mechanism to reimplement~\cite{xue2019synthetic}. For fair comparison, we reimplement~\cite{xue2019synthetic} using same pool of synthetic images.}
\label{tb:cervical}
\vspace{-5pt}
\end{table*}

\begin{table*}[!hbtp]
\begin{center}
\begin{tabular}{|l|c|c|c|c|}
\hline
{} &        CIFAR (10\%) &        CIFAR (25\%) &        CIFAR (50\%) &        CIFAR (75\%) \\
\hline
Baseline Model~\cite{he2016deep}             &  0.539 $\pm$ 0.006 &  0.655 $\pm$ 0.003 &  0.717 $\pm$ 0.010 &  0.763 $\pm$ 0.003 \
\\
\hline
\enspace + Traditional Augmentation &  0.540 $\pm$ 0.010 &  0.669 $\pm$ 0.013 &  0.735 $\pm$ 0.005 &  0.768 $\pm$ 0.006 \\
\hline
\enspace + GAN Augmentation, r=0.5     &  0.540 $\pm$ 0.003 &  0.676 $\pm$ 0.004 &  0.738 $\pm$ 0.002 &  0.768 $\pm$ 0.003 \\
\hline
\enspace + \text{Selective Augmentation, r=0.5}     & \textbf{0.578 $\pm$ 0.004} &  \textbf{0.692 $\pm$ 0.004} &  \textbf{0.745 $\pm$ 0.002} &  \textbf{0.776 $\pm$ 0.003} \\
\hline
\end{tabular}
\end{center}
\caption{The accuracy score of baseline and augmentation models using subsets of CIFAR10 with different percentage as the training set. All evaluations are done on the whole test set of CIFAR10.}
\label{tb:cifar_acc}
\vspace{-5pt}
\end{table*}

 In Table~\ref{tb:cervical}, we compare quantitative results with different baselines along with an ablation study of augmentation ratio $r$. In all experiments, we use the same backbone ResNet34 classifier for fair comparison. For our cGAN model without image selection, although the synthetic images are already realistic, classification performance is actually downgraded compared to baseline model without any augmentation and with traditional augmentations including horizontal flipping and color jittering. This result again proves that blindly expanding training set with synthetic images, even if they are realistic enough, could be harmful rather than helpful. For our model with image selection, we first compare with the prior work~\cite{xue2019synthetic}. Note that we use an improved cGAN model and achieve better synthetic images than in~\cite{xue2019synthetic} (as shown in the Figure~\ref{fig:cervical}). Moreover, \cite{xue2019synthetic} only uses a single feature extractor with a single run and does not consider the confidence of assigned labels. We use the same cGAN model which generates more visually appealing results than in~\cite{xue2019synthetic} to validate the effectiveness of our new image selection method. Under the same ratio $r=0.5$, our image selection model improves the classification result of~\cite{xue2019synthetic} by a large margin. To provide some insight on how the choice of $r$ affects the augmentation result, we also conduct an ablation study using different values of $r$. One can see that large value of $r$ compromises the advantage of synthetic augmentation. One possible explanation is that given the same number of generated images, the proportion of images which can provide meaningful features for data augmentation is limited. Thus, an image selection process is indispensable. Since $r=0.5$ achieves best performance, we keep this value in the natural image classification experiments.

In Table~\ref{tb:cifar_acc} and Table~\ref{tb:svhn_f1}, our model also significantly outperforms all baseline models, especially when the number of available training data is very small (such as in 10\% CIFAR10). By conducting experiments on two standard natural image benchmarks, we prove that our proposed pipeline is general and can be applied to various types of tasks.

\begin{table}[t]
\begin{center}
\resizebox{0.99\columnwidth}{!}
{
\begin{tabular}{|c|c|c|c|}
\hline
Baseline &    \enspace + Traditional & \enspace + GAN   &  \enspace + Selective \\
Model~\cite{he2016deep}  &   Aug.   & Aug., r=0.5  &   Aug., r=0.5  \\ 
\hline
0.918 $\pm$ 0.012 &  0.927 $\pm$ 0.001 & 0.931 $\pm$ 0.001  &  \textbf{0.934 $\pm$ 0.001} \\
\hline
\end{tabular}
}
\end{center}
\caption{The accuracy of baseline and augmentation models using 25\% of SVHN as the training set.}
\label{tb:svhn_f1}
\vspace{-5pt}
\end{table}

\section{Discussion}
While our proposed selective synthetic augmentation significantly outperforms all baseline models, partial credits should go to the high-fidelity images generated by our proposed cGAN and the state-of-the-art BigGAN. Besides the visual quality of images, the diversity of images plays an important role in synthetic augmentation. Since synthetic augmentation is imperative in scenarios with very scarce training samples, combining our pipeline with a GAN model that can learn from limited data~\cite{wang2018transferring, noguchi2019image} would further improve the generality of our method. As we provide a solution to assure the synthetic image quality during augmentation, there is room for improvement in the selection mechanism. More advanced methods for model selection and image selection, such as an end-to-end method, will be investigated in future works.

\section{Conclusion}
In this paper, we propose a synthetic augmentation method with quality assurance. By selectively adding high fidelity samples generated by cGANs into the original dataset, our model remarkably boosts the classification performance of baseline models. Experiments on both medical image and natural image datasets demonstrate the effectiveness and generality of our method.

{\small
\bibliographystyle{ieee_fullname}
\bibliography{egbib}
}

\end{document}